\documentclass[]{utexas}
\usepackage[utf8]{inputenc} 
\usepackage[T1]{fontenc}    
\usepackage{hyperref}       
\usepackage{url}            
\usepackage{booktabs}       
\usepackage{amsfonts}       
\usepackage{nicefrac}       
\usepackage{microtype}      
\usepackage{xcolor}         

\usepackage{amsmath}
\usepackage{enumerate} 
\usepackage{algorithm}
\usepackage[noend]{algpseudocode}
\usepackage{amsfonts}
\usepackage{amsthm}
\usepackage{newtxtt}
\usepackage{cleveref}
\usepackage{diagbox} 
\usepackage{colortbl}
\usepackage{nicematrix}
\usepackage{subcaption}
\usepackage{multirow}

\usepackage{tcolorbox}
\usepackage[framemethod=tikz]{mdframed}
\usepackage{amssymb}
\usepackage{xspace}
\usepackage{wrapfig}
\usepackage{adjustbox}
\usepackage{tabularx}
\usepackage{mathtools}
\usepackage{xcolor}         

\usepackage{array}
\usepackage[table]{xcolor}
\usepackage{colortbl}
\usepackage[normalem]{ulem} 
\usepackage{makecell}  
\usepackage{graphicx}

\usepackage[table]{xcolor}
\usepackage{array}
\usepackage{booktabs}
\usepackage{pifont}
\definecolor{hdrblue}{HTML}{0E7490}
\definecolor{sdrgray}{HTML}{6B7280}
\definecolor{mint}{HTML}{10B981}
\definecolor{amber}{HTML}{F59E0B}
\definecolor{rose}{HTML}{F43F5E}
\definecolor{violet}{HTML}{7C3AED}
\definecolor{sky}{HTML}{0EA5E9}
\definecolor{slate}{HTML}{1F2937}
\definecolor{panelbg}{HTML}{F8FAFC}
\usepackage{pifont}
\newcommand{\cmark}{\ding{51}}  
\newcommand{\xmark}{\ding{55}}  

\usepackage{xcolor}

\definecolor{OursGreen}{RGB}{232,245,233}      
\definecolor{ColGray}{RGB}{245,245,245}        
\definecolor{HeaderGray}{RGB}{250,250,250}

\renewcommand{\arraystretch}{1.10}

\theoremstyle{plain}

\theoremstyle{definition}

\theoremstyle{remark}

\usepackage[textsize=tiny]{todonotes}

\usepackage{tikz}

\title{Seeing Beyond 8bits: Subjective and Objective Quality Assessment of HDR-UGC Videos}

\author[1]{Shreshth Saini}
\author[1]{Bowen Chen}
\author[2]{Neil Birkbeck}
\author[2]{Yilin Wang}
\author[2]{Balu Adsumilli}
\author[1]{Alan C. Bovik}

\affiliation[1]{Laboratory for Image and Video Engineering (LIVE), UT Austin.}
\affiliation[2]{Google/YouTube}

\footnotetext{\textdagger LIVE-YouTube Beyond8Bits Dataset.}

\abstract{
High Dynamic Range (HDR) user-generated (UGC) videos are rapidly proliferating across social platforms, yet most perceptual video quality assessment (VQA) systems remain tailored to Standard Dynamic Range (SDR). HDR’s higher bit depth, wide color gamut, and elevated luminance range expose distortions such as near-black crushing, highlight clipping, banding, and exposure flicker that amplify UGC artifacts and challenge SDR models. To catalyze progress, we curate \textbf{Beyond8Bits}, a large-scale subjective dataset of $\sim$44K videos from 6.5K sources with $>$1.5M crowd ratings, spanning diverse scenes, capture conditions, and compression settings. We further introduce \textbf{HDR-Q}, the first Multimodal Large Language Model (MLLM) for HDR-UGC VQA. We propose (i) a novel HDR-aware vision encoder to produce HDR-sensitive embeddings, and (ii) HDR-Aware Policy Optimization (HAPO), an RL finetuning framework that anchors reasoning to HDR cues. HAPO augments GRPO via an HDR–SDR contrastive KL that encourages token reliance on HDR inputs and a gaussian weighted regression reward for fine-grained MOS calibration. Across \textbf{Beyond8Bits} and public HDR-VQA benchmarks, \textbf{HDR-Q} delivers state-of-the-art performance. 
}

\date{\today}
\page{\url{https://shreshthsaini.github.io/Beyond8Bits}}
\correspondence{Shreshth Saini at \email{saini.2@utexas.edu}}

\begin{document}


\newcommand*{\vertbar}{\rule[-0.25ex]{0.5pt}{1.5ex}}
\newcommand*{\horzbar}{\rule[.5ex]{2.5ex}{0.5pt}}
\newcommand{\dd}{\mathrm{d}}
\newcommand{\tkernel}{p}
\newcommand{\action}[2]{\left \langle #1, #2\right \rangle }
\newcommand{\bell}{\mathrm{b}}
\newcommand{\norm}[1]
{\left\Vert#1\right\Vert}
\newcommand{\Norm}[1]{\lvert \! \lvert \! \lvert #1 \rvert \! \rvert \! \rvert}
\newcommand{\abs}[1]{\left\vert#1\right\vert}
\newcommand{\babs}[1]{\Big \vert#1 \Big \vert}
\newcommand{\set}[1]{\left\{#1\right\}}
\newcommand{\parr}[1]{\left (#1\right )}
\newcommand{\brac}[1]{\left [#1\right ]}
\newcommand{\ip}[1]{\left \langle #1 \right \rangle }
\newcommand{\Real}{\mathbb R}
\newcommand{\Nat}{\mathbb N}
\newcommand{\Complex}{\mathbb C}
\newcommand{\eps}{\varepsilon}
\newcommand{\too}{\rightarrow}
\newcommand{\bbar}[1]{\overline{#1}}
\newcommand{\wt}[1]{\widetilde{#1}} 
\newcommand{\wh}[1]{\widehat{#1}} 
\newcommand{\diag}{\textrm{diag}} 
\newcommand{\dist}{d} 
\newcommand{\divv}{\mathrm{div}} 
\newcommand{\vol}{\mathrm{vol}} 
\newcommand{\snr}{\mathrm{snr}}
\newcommand{\logsnr}{\rho}
\newcommand{\trace}{\textrm{tr}} 
\def \bfi{\textbf{\footnotesize{i}}} 
\newcommand{\one}{\mathbf{1}}
\newcommand{\zero}{\mathbf{0}}
\newcommand{\vcc}[1]{\mathrm{vec}(#1)}
\newcommand{\mat}[1]{\bm{[} #1 \bm{]}}
\newcommand{\defe}{\coloneqq}

\definecolor{mygray}{gray}{0.95}
\newcommand{\CM}{\scriptscriptstyle \text{CM}}
\newcommand{\M}{\scriptscriptstyle \text{M}}
\newcommand{\CFM}{\scriptscriptstyle \text{CFM}}
\newcommand{\FM}{\scriptscriptstyle \text{FM}}
\newcommand{\RFM}{\scriptscriptstyle \text{RFM}}
\newcommand{\RCFM}{\scriptscriptstyle \text{RCFM}}
\newcommand{\DFM}{\scriptscriptstyle \text{DFM}}
\newcommand{\CDFM}{\scriptscriptstyle \text{CDFM}}
\newcommand{\GM}{\scriptscriptstyle \text{GM}}
\newcommand{\DSM}{\scriptscriptstyle \text{DSM}}
\newcommand{\CGM}{\scriptscriptstyle \text{CGM}}
\newcommand{\SM}{\scriptscriptstyle \text{SM}}
\newcommand{\NM}{\scriptscriptstyle \text{NM}}
\newcommand{\mask}{\texttt{m}} 
\newcommand{\ignore}{\texttt{i}} 

\def \etal{{et al}.}
\newcommand*{\eg}{{\it e.g.}\@\xspace}
\newcommand*{\ie}{{\it i.e.}\@\xspace}

\makeatletter
\newtheorem*{rep@theorem}{\rep@title}
\newcommand{\newreptheorem}[2]{%
\newenvironment{rep#1}[1]{%
 \def\rep@title{\textbf{#2} \ref{##1}}%
 \begin{rep@theorem}}%
 {\end{rep@theorem}}}
\makeatother


\newreptheorem{theorem}{Theorem}
\newreptheorem{proposition}{Proposition}
\newreptheorem{lemma}{Lemma}
\newreptheorem{corollary}{Corollary}


\newcommand{\figleft}{{\em (Left)}}
\newcommand{\figcenter}{{\em (Center)}}
\newcommand{\figright}{{\em (Right)}}
\newcommand{\figtop}{{\em (Top)}}
\newcommand{\figbottom}{{\em (Bottom)}}
\newcommand{\captiona}{{\em (a)}}
\newcommand{\captionb}{{\em (b)}}
\newcommand{\captionc}{{\em (c)}}
\newcommand{\captiond}{{\em (d)}}

\newcommand{\newterm}[1]{{\bf #1}}

\def\figref#1{figure~\ref{#1}}
\def\Figref#1{Figure~\ref{#1}}
\def\twofigref#1#2{figures \ref{#1} and \ref{#2}}
\def\quadfigref#1#2#3#4{figures \ref{#1}, \ref{#2}, \ref{#3} and \ref{#4}}
\def\secref#1{section~\ref{#1}}
\def\Secref#1{Section~\ref{#1}}
\def\twosecrefs#1#2{sections \ref{#1} and \ref{#2}}
\def\secrefs#1#2#3{sections \ref{#1}, \ref{#2} and \ref{#3}}
\def\eqref#1{equation~\ref{#1}}
\def\Eqref#1{Equation~\ref{#1}}
\def\plaineqref#1{\ref{#1}}
\def\chapref#1{chapter~\ref{#1}}
\def\Chapref#1{Chapter~\ref{#1}}
\def\rangechapref#1#2{chapters\ref{#1}--\ref{#2}}
\def\algref#1{algorithm~\ref{#1}}
\def\Algref#1{Algorithm~\ref{#1}}
\def\twoalgref#1#2{algorithms \ref{#1} and \ref{#2}}
\def\Twoalgref#1#2{Algorithms \ref{#1} and \ref{#2}}
\def\partref#1{part~\ref{#1}}
\def\Partref#1{Part~\ref{#1}}
\def\twopartref#1#2{parts \ref{#1} and \ref{#2}}

\def\ceil#1{\lceil #1 \rceil}
\def\floor#1{\lfloor #1 \rfloor}
\def\1{\bm{1}}
\newcommand{\train}{\mathcal{D}}
\newcommand{\valid}{\mathcal{D_{\mathrm{valid}}}}
\newcommand{\test}{\mathcal{D_{\mathrm{test}}}}

\def\eps{{\epsilon}}




\def\reta{{\textnormal{$\eta$}}}
\def\ra{{\textnormal{a}}}
\def\rb{{\textnormal{b}}}
\def\rc{{\textnormal{c}}}
\def\rd{{\textnormal{d}}}
\def\re{{\textnormal{e}}}
\def\rf{{\textnormal{f}}}
\def\rg{{\textnormal{g}}}
\def\rh{{\textnormal{h}}}
\def\ri{{\textnormal{i}}}
\def\rj{{\textnormal{j}}}
\def\rk{{\textnormal{k}}}
\def\rl{{\textnormal{l}}}
\def\rn{{\textnormal{n}}}
\def\ro{{\textnormal{o}}}
\def\rp{{\textnormal{p}}}
\def\rq{{\textnormal{q}}}
\def\rr{{\textnormal{r}}}
\def\rs{{\textnormal{s}}}
\def\rt{{\textnormal{t}}}
\def\ru{{\textnormal{u}}}
\def\rv{{\textnormal{v}}}
\def\rw{{\textnormal{w}}}
\def\rx{{\textnormal{x}}}
\def\ry{{\textnormal{y}}}
\def\rz{{\textnormal{z}}}

\def\rvepsilon{{\mathbf{\epsilon}}}
\def\rvtheta{{\mathbf{\theta}}}
\def\rva{{\mathbf{a}}}
\def\rvb{{\mathbf{b}}}
\def\rvc{{\mathbf{c}}}
\def\rvd{{\mathbf{d}}}
\def\rve{{\mathbf{e}}}
\def\rvf{{\mathbf{f}}}
\def\rvg{{\mathbf{g}}}
\def\rvh{{\mathbf{h}}}
\def\rvu{{\mathbf{i}}}
\def\rvj{{\mathbf{j}}}
\def\rvk{{\mathbf{k}}}
\def\rvl{{\mathbf{l}}}
\def\rvm{{\mathbf{m}}}
\def\rvn{{\mathbf{n}}}
\def\rvo{{\mathbf{o}}}
\def\rvp{{\mathbf{p}}}
\def\rvq{{\mathbf{q}}}
\def\rvr{{\mathbf{r}}}
\def\rvs{{\mathbf{s}}}
\def\rvt{{\mathbf{t}}}
\def\rvu{{\mathbf{u}}}
\def\rvv{{\mathbf{v}}}
\def\rvw{{\mathbf{w}}}
\def\rvx{{\mathbf{x}}}
\def\rvy{{\mathbf{y}}}
\def\rvz{{\mathbf{z}}}

\def\erva{{\textnormal{a}}}
\def\ervb{{\textnormal{b}}}
\def\ervc{{\textnormal{c}}}
\def\ervd{{\textnormal{d}}}
\def\erve{{\textnormal{e}}}
\def\ervf{{\textnormal{f}}}
\def\ervg{{\textnormal{g}}}
\def\ervh{{\textnormal{h}}}
\def\ervi{{\textnormal{i}}}
\def\ervj{{\textnormal{j}}}
\def\ervk{{\textnormal{k}}}
\def\ervl{{\textnormal{l}}}
\def\ervm{{\textnormal{m}}}
\def\ervn{{\textnormal{n}}}
\def\ervo{{\textnormal{o}}}
\def\ervp{{\textnormal{p}}}
\def\ervq{{\textnormal{q}}}
\def\ervr{{\textnormal{r}}}
\def\ervs{{\textnormal{s}}}
\def\ervt{{\textnormal{t}}}
\def\ervu{{\textnormal{u}}}
\def\ervv{{\textnormal{v}}}
\def\ervw{{\textnormal{w}}}
\def\ervx{{\textnormal{x}}}
\def\ervy{{\textnormal{y}}}
\def\ervz{{\textnormal{z}}}

\def\rmA{{\mathbf{A}}}
\def\rmB{{\mathbf{B}}}
\def\rmC{{\mathbf{C}}}
\def\rmD{{\mathbf{D}}}
\def\rmE{{\mathbf{E}}}
\def\rmF{{\mathbf{F}}}
\def\rmG{{\mathbf{G}}}
\def\rmH{{\mathbf{H}}}
\def\rmI{{\mathbf{I}}}
\def\rmJ{{\mathbf{J}}}
\def\rmK{{\mathbf{K}}}
\def\rmL{{\mathbf{L}}}
\def\rmM{{\mathbf{M}}}
\def\rmN{{\mathbf{N}}}
\def\rmO{{\mathbf{O}}}
\def\rmP{{\mathbf{P}}}
\def\rmQ{{\mathbf{Q}}}
\def\rmR{{\mathbf{R}}}
\def\rmS{{\mathbf{S}}}
\def\rmT{{\mathbf{T}}}
\def\rmU{{\mathbf{U}}}
\def\rmV{{\mathbf{V}}}
\def\rmW{{\mathbf{W}}}
\def\rmX{{\mathbf{X}}}
\def\rmY{{\mathbf{Y}}}
\def\rmZ{{\mathbf{Z}}}

\def\ermA{{\textnormal{A}}}
\def\ermB{{\textnormal{B}}}
\def\ermC{{\textnormal{C}}}
\def\ermD{{\textnormal{D}}}
\def\ermE{{\textnormal{E}}}
\def\ermF{{\textnormal{F}}}
\def\ermG{{\textnormal{G}}}
\def\ermH{{\textnormal{H}}}
\def\ermI{{\textnormal{I}}}
\def\ermJ{{\textnormal{J}}}
\def\ermK{{\textnormal{K}}}
\def\ermL{{\textnormal{L}}}
\def\ermM{{\textnormal{M}}}
\def\ermN{{\textnormal{N}}}
\def\ermO{{\textnormal{O}}}
\def\ermP{{\textnormal{P}}}
\def\ermQ{{\textnormal{Q}}}
\def\ermR{{\textnormal{R}}}
\def\ermS{{\textnormal{S}}}
\def\ermT{{\textnormal{T}}}
\def\ermU{{\textnormal{U}}}
\def\ermV{{\textnormal{V}}}
\def\ermW{{\textnormal{W}}}
\def\ermX{{\textnormal{X}}}
\def\ermY{{\textnormal{Y}}}
\def\ermZ{{\textnormal{Z}}}

\def\vzero{{\bm{0}}}
\def\vone{{\bm{1}}}
\def\vmu{{\bm{\mu}}}
\def\vtheta{{\bm{\theta}}}
\def\va{{\bm{a}}}
\def\vb{{\bm{b}}}
\def\vc{{\bm{c}}}
\def\vd{{\bm{d}}}
\def\ve{{\bm{e}}}
\def\vf{{\bm{f}}}
\def\vg{{\bm{g}}}
\def\vh{{\bm{h}}}
\def\vi{{\bm{i}}}
\def\vj{{\bm{j}}}
\def\vk{{\bm{k}}}
\def\vl{{\bm{l}}}
\def\vm{{\bm{m}}}
\def\vn{{\bm{n}}}
\def\vo{{\bm{o}}}
\def\vp{{\bm{p}}}
\def\vq{{\bm{q}}}
\def\vr{{\bm{r}}}
\def\vs{{\bm{s}}}
\def\vt{{\bm{t}}}
\def\vu{{\bm{u}}}
\def\vv{{\bm{v}}}
\def\vw{{\bm{w}}}
\def\vx{{\bm{x}}}
\def\vy{{\bm{y}}}
\def\vz{{\bm{z}}}
\def\valpha{{\bm{\alpha}}}

\def\evalpha{{\alpha}}
\def\evbeta{{\beta}}
\def\evepsilon{{\epsilon}}
\def\evlambda{{\lambda}}
\def\evomega{{\omega}}
\def\evmu{{\mu}}
\def\evpsi{{\psi}}
\def\evsigma{{\sigma}}
\def\evtheta{{\theta}}
\def\eva{{a}}
\def\evb{{b}}
\def\evc{{c}}
\def\evd{{d}}
\def\eve{{e}}
\def\evf{{f}}
\def\evg{{g}}
\def\evh{{h}}
\def\evi{{i}}
\def\evj{{j}}
\def\evk{{k}}
\def\evl{{l}}
\def\evm{{m}}
\def\evn{{n}}
\def\evo{{o}}
\def\evp{{p}}
\def\evq{{q}}
\def\evr{{r}}
\def\evs{{s}}
\def\evt{{t}}
\def\evu{{u}}
\def\evv{{v}}
\def\evw{{w}}
\def\evx{{x}}
\def\evy{{y}}
\def\evz{{z}}

\def\mA{{\bm{A}}}
\def\mB{{\bm{B}}}
\def\mC{{\bm{C}}}
\def\mD{{\bm{D}}}
\def\mE{{\bm{E}}}
\def\mF{{\bm{F}}}
\def\mG{{\bm{G}}}
\def\mH{{\bm{H}}}
\def\mI{{\bm{I}}}
\def\mJ{{\bm{J}}}
\def\mK{{\bm{K}}}
\def\mL{{\bm{L}}}
\def\mM{{\bm{M}}}
\def\mN{{\bm{N}}}
\def\mO{{\bm{O}}}
\def\mP{{\bm{P}}}
\def\mQ{{\bm{Q}}}
\def\mR{{\bm{R}}}
\def\mS{{\bm{S}}}
\def\mT{{\bm{T}}}
\def\mU{{\bm{U}}}
\def\mV{{\bm{V}}}
\def\mW{{\bm{W}}}
\def\mX{{\bm{X}}}
\def\mY{{\bm{Y}}}
\def\mZ{{\bm{Z}}}
\def\mBeta{{\bm{\beta}}}
\def\mPhi{{\bm{\Phi}}}
\def\mphi{\bm{\phi}}
\def\mPsi{{\bm{\Psi}}}
\def\mpsi{\bm{\psi}}
\def\mLambda{{\bm{\Lambda}}}
\def\mSigma{{\bm{\Sigma}}}

\newcommand{\tens}[1]{\bm{\mathsfit{#1}}}
\def\tA{{\tens{A}}}
\def\tB{{\tens{B}}}
\def\tC{{\tens{C}}}
\def\tD{{\tens{D}}}
\def\tE{{\tens{E}}}
\def\tF{{\tens{F}}}
\def\tG{{\tens{G}}}
\def\tH{{\tens{H}}}
\def\tI{{\tens{I}}}
\def\tJ{{\tens{J}}}
\def\tK{{\tens{K}}}
\def\tL{{\tens{L}}}
\def\tM{{\tens{M}}}
\def\tN{{\tens{N}}}
\def\tO{{\tens{O}}}
\def\tP{{\tens{P}}}
\def\tQ{{\tens{Q}}}
\def\tR{{\tens{R}}}
\def\tS{{\tens{S}}}
\def\tT{{\tens{T}}}
\def\tU{{\tens{U}}}
\def\tV{{\tens{V}}}
\def\tW{{\tens{W}}}
\def\tX{{\tens{X}}}
\def\tY{{\tens{Y}}}
\def\tZ{{\tens{Z}}}

\def\gA{{\mathcal{A}}}
\def\gB{{\mathcal{B}}}
\def\gC{{\mathcal{C}}}
\def\gD{{\mathcal{D}}}
\def\gE{{\mathcal{E}}}
\def\gF{{\mathcal{F}}}
\def\gG{{\mathcal{G}}}
\def\gH{{\mathcal{H}}}
\def\gI{{\mathcal{I}}}
\def\gJ{{\mathcal{J}}}
\def\gK{{\mathcal{K}}}
\def\gL{{\mathcal{L}}}
\def\gM{{\mathcal{M}}}
\def\gN{{\mathcal{N}}}
\def\gO{{\mathcal{O}}}
\def\gP{{\mathcal{P}}}
\def\gQ{{\mathcal{Q}}}
\def\gR{{\mathcal{R}}}
\def\gS{{\mathcal{S}}}
\def\gT{{\mathcal{T}}}
\def\gU{{\mathcal{U}}}
\def\gV{{\mathcal{V}}}
\def\gW{{\mathcal{W}}}
\def\gX{{\mathcal{X}}}
\def\gY{{\mathcal{Y}}}
\def\gZ{{\mathcal{Z}}}

\def\sA{{\mathbb{A}}}
\def\sB{{\mathbb{B}}}
\def\sC{{\mathbb{C}}}
\def\sD{{\mathbb{D}}}
\def\sF{{\mathbb{F}}}
\def\sG{{\mathbb{G}}}
\def\sH{{\mathbb{H}}}
\def\sI{{\mathbb{I}}}
\def\sJ{{\mathbb{J}}}
\def\sK{{\mathbb{K}}}
\def\sL{{\mathbb{L}}}
\def\sM{{\mathbb{M}}}
\def\sN{{\mathbb{N}}}
\def\sO{{\mathbb{O}}}
\def\sP{{\mathbb{P}}}
\def\sQ{{\mathbb{Q}}}
\def\sR{{\mathbb{R}}}
\def\sS{{\mathbb{S}}}
\def\sT{{\mathbb{T}}}
\def\sU{{\mathbb{U}}}
\def\sV{{\mathbb{V}}}
\def\sW{{\mathbb{W}}}
\def\sX{{\mathbb{X}}}
\def\sY{{\mathbb{Y}}}
\def\sZ{{\mathbb{Z}}}

\def\emLambda{{\Lambda}}
\def\emA{{A}}
\def\emB{{B}}
\def\emC{{C}}
\def\emD{{D}}
\def\emE{{E}}
\def\emF{{F}}
\def\emG{{G}}
\def\emH{{H}}
\def\emI{{I}}
\def\emJ{{J}}
\def\emK{{K}}
\def\emL{{L}}
\def\emM{{M}}
\def\emN{{N}}
\def\emO{{O}}
\def\emP{{P}}
\def\emQ{{Q}}
\def\emR{{R}}
\def\emS{{S}}
\def\emT{{T}}
\def\emU{{U}}
\def\emV{{V}}
\def\emW{{W}}
\def\emX{{X}}
\def\emY{{Y}}
\def\emZ{{Z}}
\def\emSigma{{\Sigma}}

\newcommand{\etens}[1]{\mathsfit{#1}}
\def\etLambda{{\etens{\Lambda}}}
\def\etA{{\etens{A}}}
\def\etB{{\etens{B}}}
\def\etC{{\etens{C}}}
\def\etD{{\etens{D}}}
\def\etE{{\etens{E}}}
\def\etF{{\etens{F}}}
\def\etG{{\etens{G}}}
\def\etH{{\etens{H}}}
\def\etI{{\etens{I}}}
\def\etJ{{\etens{J}}}
\def\etK{{\etens{K}}}
\def\etL{{\etens{L}}}
\def\etM{{\etens{M}}}
\def\etN{{\etens{N}}}
\def\etO{{\etens{O}}}
\def\etP{{\etens{P}}}
\def\etQ{{\etens{Q}}}
\def\etR{{\etens{R}}}
\def\etS{{\etens{S}}}
\def\etT{{\etens{T}}}
\def\etU{{\etens{U}}}
\def\etV{{\etens{V}}}
\def\etW{{\etens{W}}}
\def\etX{{\etens{X}}}
\def\etY{{\etens{Y}}}
\def\etZ{{\etens{Z}}}

\newcommand{\pdata}{p_{\rm{data}}}
\newcommand{\ptrain}{\hat{p}_{\rm{data}}}
\newcommand{\Ptrain}{\hat{P}_{\rm{data}}}
\newcommand{\pmodel}{p_{\rm{model}}}
\newcommand{\Pmodel}{P_{\rm{model}}}
\newcommand{\ptildemodel}{\tilde{p}_{\rm{model}}}
\newcommand{\pencode}{p_{\rm{encoder}}}
\newcommand{\pdecode}{p_{\rm{decoder}}}
\newcommand{\precons}{p_{\rm{reconstruct}}}

\newcommand{\laplace}{\mathrm{Laplace}} 

\newcommand{\E}{\mathbb{E}}
\newcommand{\Ls}{\mathcal{L}}
\newcommand{\R}{\mathbb{R}}
\newcommand{\emp}{\tilde{p}}
\newcommand{\lr}{\alpha}
\newcommand{\reg}{\lambda}
\newcommand{\rect}{\mathrm{rectifier}}
\newcommand{\softmax}{\mathrm{softmax}}
\newcommand{\sigmoid}{\sigma}
\newcommand{\softplus}{\zeta}
\newcommand{\KL}{D_{\mathrm{KL}}}
\newcommand{\Var}{\mathrm{Var}}
\newcommand{\standarderror}{\mathrm{SE}}
\newcommand{\Cov}{\mathrm{Cov}}
\newcommand{\normlzero}{L^0}
\newcommand{\normlone}{L^1}
\newcommand{\normltwo}{L^2}
\newcommand{\normlp}{L^p}
\newcommand{\normmax}{L^\infty}

\newcommand{\parents}{Pa} 

\let\ab\allowbreak

\newcolumntype{C}[1]{>{\Centering}m{#1}}
\renewcommand\tabularxcolumn[1]{C{#1}}
\newcolumntype{Z}[1]{>{\Left}m{#1}}
\renewcommand\tabularxcolumn[1]{Z{#1}}

\maketitle




\section{Introduction}
\label{introduction}
\begin{figure*}[htbp]
    \centering
    \includegraphics[width=\textwidth]{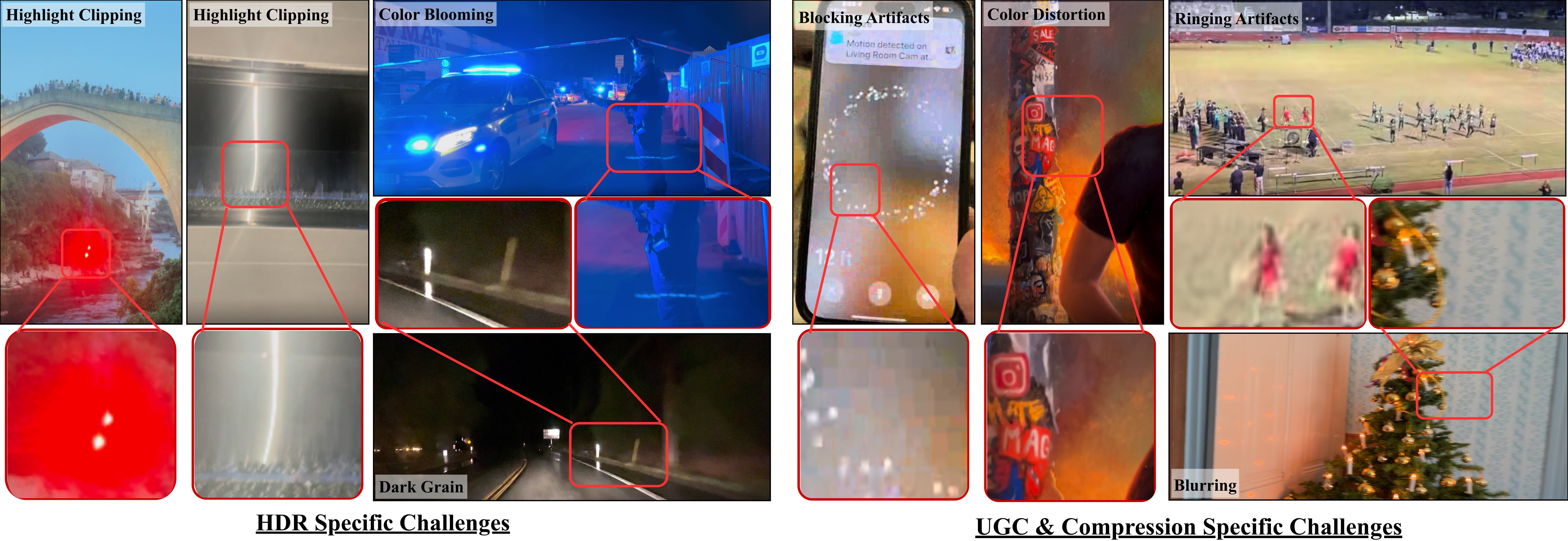}
    \caption{Typical challenges in HDR-UGC videos, including HDR-specific issues (e.g., highlight clipping, color blooming, dark grain) and UGC/compression-related artifacts (e.g., color distortion, blocking, ringing, blurring).}
    \label{fig:artifacts}
\end{figure*}

The digital media ecosystem has been transformed by the explosive growth of user-generated content (UGC) on platforms such as YouTube, TikTok, and Instagram~\citep{99firms_facebook,omnicore_tiktok,mohsin2020youtube}. In parallel, High Dynamic Range (HDR) video has become mainstream, offering higher bit depth, wider color gamut, and extended luminance range compared to Standard Dynamic Range (SDR)~\citep{HDR-SDR_bwchen,LIVE-HDR,HIDRO-VQA, chug}. These characteristics enhance perceptual realism but also accentuate distortions that are less visible in SDR such as near-black crushing, highlight clipping, banding, and exposure flicker often compounded by compression or capture artifacts common in UGC (Fig.~\ref{fig:artifacts}). As a result, evaluating the perceptual quality of HDR-UGC content remains an open and underexplored challenge. Existing Video Quality Assessment (VQA) models struggle in this regime. Methods trained on professionally generated HDR datasets~\citep{LIVE-HDR,HDR-SDR_bwchen} or SDR-UGC videos~\citep{KVQ,MDVQA} fail to generalize to the heterogeneous capture conditions, device variations, and uncontrolled distortions of real-world HDR-UGC. Moreover, current HDR subjective datasets are small and limited in scope, focusing primarily on synthetic distortions or curated professional content~\citep{sfv+hdr,HDRorSDR}. This lack of large-scale, real-world HDR-UGC annotations represents a major obstacle to developing models that align with human perceptual judgments.

\begin{figure*}[htbp]
    \centering
    \includegraphics[width=0.88\textwidth]{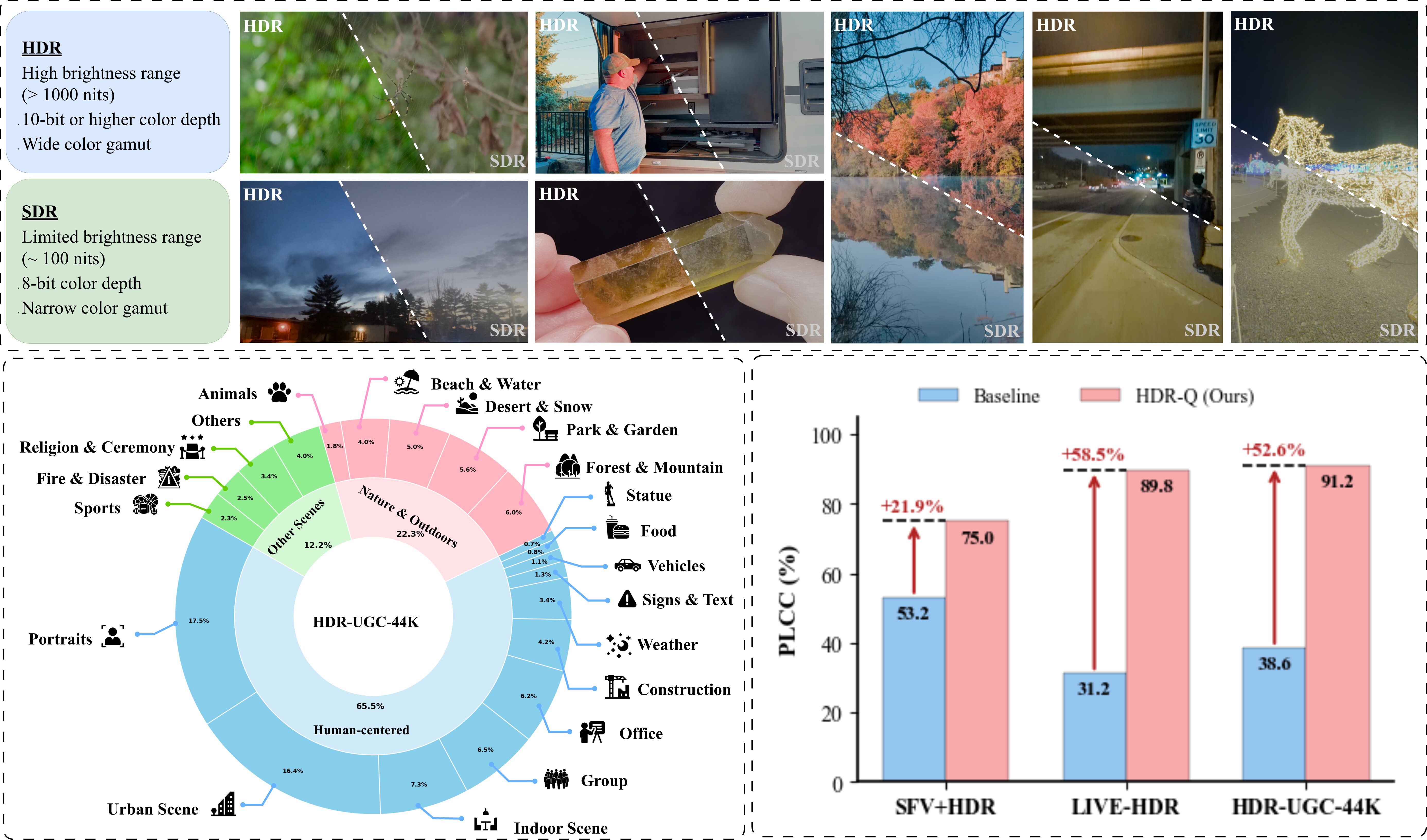}
    \caption{
    \textbf{Overview of our dataset and performance evaluation.} Top: Example comparisons between HDR and SDR frames, illustrating differences in brightness range, color depth, and visual detail across diverse scenes. Bottom-left: The distribution of video categories in the \textbf{Beyond8Bits} dataset, covering human-centered content, nature \& outdoor scenes, and various other real-world scenarios. Bottom-right: Performance comparison between our proposed HDR-Q model and baseline methods on three datasets, where HDR-Q achieves significant improvements in PLCC.
    }  
    \label{fig:overview}
\end{figure*}

Meanwhile, multimodal large language models (MLLMs) have emerged as powerful reasoning systems that unify perception and language, showing promise for explainable image and video quality assessment~\citep{qalign,qinstruct,coinstruct,deqa,depictqa,visualquality,qinsight}. However, their direct application to HDR-UGC VQA faces key obstacles:  
(i) standard visual encoders are pre-trained on SDR data and fail to capture HDR-specific cues;  
(ii) obtaining accurate, continuous MOS predictions remains challenging within the next-token prediction paradigm, as discrete-level or regression-head approaches~\citep{visualquality,deqa,qalign,finevq,compare2score} lack fine-grained calibration;  
(iii) without explicit incentives, policies often neglect HDR inputs and rely on textual priors, a form of modality neglect~\citep{modalityneglect, sft-rl, papo}.

To address these challenges, we introduce \textbf{Beyond8Bits}, the first large-scale, crowdsourced subjective HDR-UGC quality dataset containing $\sim$44K videos from 6,861 diverse sources with over 1.5M human ratings. This dataset provides the necessary foundation for training and evaluating models that reflect real-world HDR perceptual phenomena. Building upon it, we propose \textcolor{hdrblue}{\textbf{HDR-Q}}, the first multimodal large language model specifically designed for HDR-UGC quality assessment. \textcolor{hdrblue}{\textbf{HDR-Q}} integrates two novel components:  
(i) an HDR-aware vision encoder that learns HDR-sensitive representations while maintaining semantic alignment, and  
(ii) HDR-Aware Policy Optimization (HAPO), a reinforcement learning framework that enforces HDR grounding through an HDR–SDR contrastive KL term, stabilizes entropy, and refines token-level credit assignment via entropy-weighted advantages. A Gaussian regression reward further enables fine-grained MOS calibration, while group-level self-rewarding improves reasoning consistency. Our contributions are:

\begin{itemize}
    \item We introduce \textbf{Beyond8Bits}, the largest subjective HDR-UGC quality dataset.
    \item We propose \textcolor{hdrblue}{\textbf{HDR-Q}}, a novel MLLM-based VQA model that combines new HDR-aware vision encoder with our novel HAPO policy, an RL finetuning paradigm tailored for perceptual reasoning under HDR conditions.
    \item Extensive experiments on \textbf{Beyond8Bits} and public HDR benchmarks demonstrate that \textcolor{hdrblue}{\textbf{HDR-Q}} achieves state-of-the-art MOS prediction and generates concise, HDR-grounded rationales, establishing a new direction for HDR-aware VQA research.
\end{itemize}

\vspace{-5pt}
\section{Related Work}
\vspace{-5pt}
\subsection{HDR-VQA: Datasets \& Models}
Early VQA datasets such as CVD2014~\citep{CVD2014}, LIVE-VQA~\citep{LIVEVQA}, LIVE-VQC~\citep{LIVE-VQC}, LSVQ~\citep{LSVQ}, MDVQA~\citep{MDVQA}, and Maxwell~\citep{maxwell} enabled both handcrafted~\citep{NIQE, BRISQUE, V-BLINDS, TLVQM, VIIDEO, ChipQA} and deep VQA models~\citep{VSFA, FASTVQA, FASTERVQA, CONTRIQUE, cover, Dover}, but remain SDR-oriented and unsuitable for HDR due to fundamental differences in luminance and tone-mapping. HDR-specific subjective datasets~\citep{azimi2021pu21, pan2018hdr, baroncini2016verification, rerabek2015subjective, athar2019perceptual} addressed this gap, though many are outdated or restricted. Recent releases such as LIVE-HDR~\citep{LIVE-HDR} (310 annotated videos) and SFV+HDR~\citep{sfv+hdr} (2{,}000 clips, 300 rated) provide more reliable benchmarks for modern HDR algorithms.
\noindent
Correspondingly, HDR-VQA models have emerged: full-reference metrics HDR-VQM~\citep{HDR-VQM}, HDR-BVQM~\citep{HDR-BVQM}, and PU21~\citep{PU21} use brightness-aware or perceptually uniform transforms but rely on references and struggle with diverse HDR distortions. Blind methods such as HDR-ChipQA~\citep{HDR-ChipQA} and HIDRO-VQA~\citep{HIDRO-VQA} extend ChipQA and CONTRIQUE~\citep{CONTRIQUE} through nonlinear luminance mappings or large-scale unlabeled HDR data. Nonetheless, existing datasets and models still fail to capture the heterogeneous degradations of HDR UGC, motivating new data and modeling strategies.

\subsection{MLLM-Based Perceptual Quality Assessment}
MLLMs have recently been explored for IQA/VQA. Benchmarks such as Q-Bench~\citep{qbench} revealed large gaps between MLLMs and human judgments, spurring instruction tuning (Q-Instruct~\citep{qinstruct}) and descriptive distortion reasoning (DepictQA, DepictQA-Wild~\citep{depictqa, depictqawild}). Comparative and ranking-based methods (Compare2Score~\citep{compare2score}, VisualQuality-R1~\citep{visualquality}) further improved human alignment, while Q-Align~\citep{qalign}, DeQA-Score~\citep{deqa}, and Q-Insight~\citep{qinsight} targeted interpretability, regression fidelity, and joint degradation reasoning. Video extensions include Q-Bench-Video~\citep{qbenchvideo} and MVQA-68K~\citep{mvqa}, which provide large-scale, multi-dimensional annotations and textual rationales for training video-aware MLLM quality evaluators.

\section{Dataset: \textbf{Beyond8Bits}}
\label{sec:dataset}
Existing HDR VQA datasets \citep{LIVE-HDR, sfv+hdr,chug, HDRorSDR, chen2025hdrsdr} 
are limited in scale, diversity, or dynamic range, and primarily focus on professionally produced content. In contrast, real-world HDR user-generated videos (HDR-UGC) exhibit a far wider range of luminance, motion, and compression characteristics, often captured under uncontrolled conditions. To bridge this gap, we introduce \textbf{Beyond8Bits}, the largest and most diverse HDR VQA dataset to date, explicitly designed for real-world HDR-UGC quality assessment.


\begin{figure*}[h!]
\centering
\includegraphics[width=\linewidth]{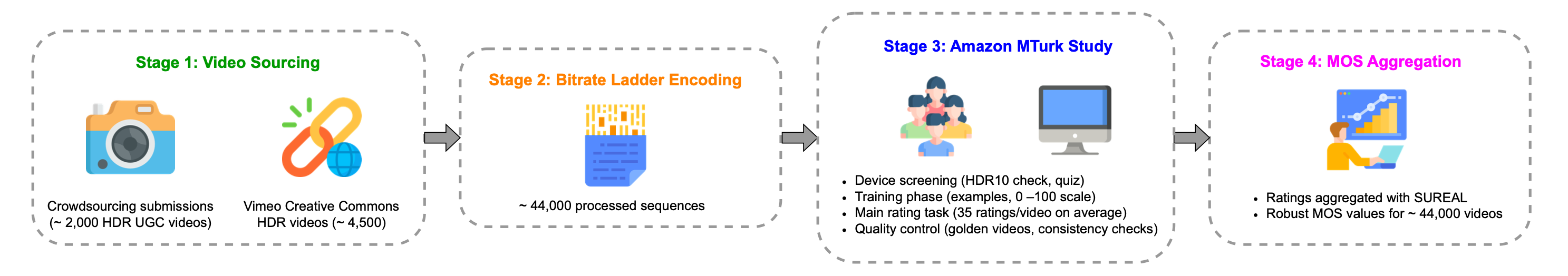}
\caption{Pipeline of \textbf{Beyond8Bits} construction.}
\label{fig:dataset_pipeline}
\end{figure*}

\subsection{Data Collection and Processing}
We collected 6{,}861 unique HDR source videos from two complementary sources: (1) a dedicated crowdsourcing campaign where users contributed HDR clips captured on consumer devices (iPhone, Pixel, Galaxy, \textit{etc.}) under research consents, contributing 2,253 videos, and (2) public HDR videos from Vimeo licensed under Creative Commons, contributing 4,608 videos. This combination ensures coverage across human-centric, natural, and low-light scenes with rich intra and inter-device variability.

Each source video was verified for HDR metadata (PQ transfer, 10-bit HEVC, BT.2020 gamut) and filtered to remove duplicates, static frames, and unsuitable content. Clips were trimmed to a maximum of 10 seconds and transcoded under a bitrate ladder simulating real-world streaming conditions~\citep{youtube_bitrate,apple_hls} at multiple resolutions (1080p–360p) and bitrates (0.2–5 Mbps), see Appendix \textcolor{red}{D}. All versions retained full HDR signaling, producing a total of $\sim$44{,}276 processed video clips, an equal mix of landscape and portrait orientations was maintained where possible. Fig.~\ref{fig:overview} visualizes the dataset composition and example content diversity.



\subsection{Subjective Quality Study}
We conducted a large-scale subjective study on Amazon Mechanical Turk (AMT), marking the first HDR large-scale crowdsourced VQA study at this scale. To ensure display fidelity, only workers with verified HDR-capable devices and browsers were admitted. Our Human Intelligence Task (HIT) design incorporated several quality control measures. Each HIT began with instructions and a qualification quiz checking for HDR display capability, and understanding of the task. 
Participants first completed a training and calibration phase with representative examples and then rated batches of clips using a continuous 0–100 likert-scale following ITU-R BT.500-14~\citep{ITU_BT500-14} guidelines. To ensure reliability, we embedded hidden quality control videos (repeats and golden-set videos with known quality ranges established in pilot studies). Strict participant screening and rejection criteria were applied based on consistency checks on control videos, display bit depth, internet speed, task completion times, and reported viewing conditions. Golden-set and repeat videos were embedded to assess intra and inter-subject consistency. Over 1.5M valid ratings were collected after rigorous quality control. Each video received on average $\sim$35 independent ratings. This large-scale design captures genuine perceptual variability under realistic HDR viewing conditions.


\subsection{MOS Aggregation}
To aggregate the subjective ratings into reliable MOS, we employed the Subjective Reliability (SUREAL) method \citep{sureal}. SUREAL provides a Maximum Likelihood Estimate (MLE) of the true video quality ($\psi_j$) by modeling individual subject ratings ($S_{ij}$) while accounting for subject bias ($\Delta_i$) and inconsistency ($\nu_i$). The model is given by:
\begin{equation}
S_{ij} = \psi_j + \Delta_i + \nu_i X, \quad X \sim \mathcal{N}(0, 1)
\end{equation}
Parameters were estimated to maximize the log-likelihood. The resulting MOS values exhibit strong inter-subject correlation (median SRCC $0.90$), confirming study reliability. 

\textbf{Beyond8Bits} spans diverse content categories (human, indoor, outdoor, night, motion-intensive), varying brightness distributions, and wide MOS coverage ($10$–$95$). Key statistics, including spatial/temporal complexity and comparisons with existing datasets, is provided in the Appendix \textcolor{red}{D}. \textbf{Beyond8Bits} provides an essential foundation for modern HDR-aware training perceptual models such as \textcolor{hdrblue}{\textbf{HDR-Q}}.

\begin{figure*}[t]
\centering
\includegraphics[width=\textwidth]{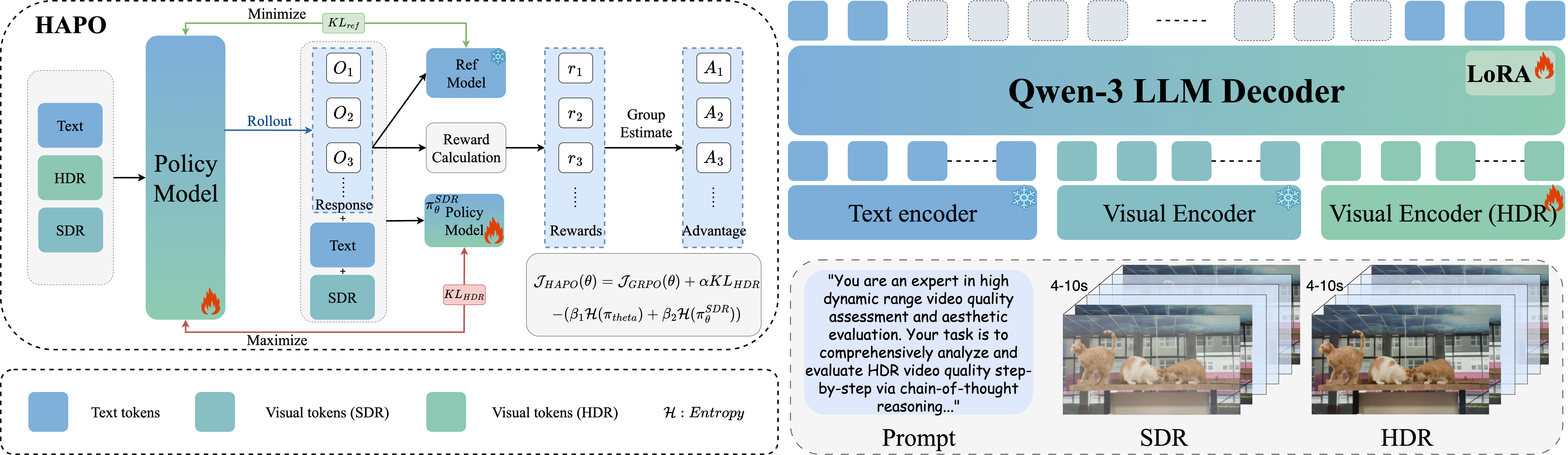}
\caption{\textbf{Overview of \textcolor{hdrblue}{HDR-Q} with \textcolor{hdrblue}{HAPO}.}
Left: HAPO compares rollouts under HDR inputs (text + SDR + HDR tokens) versus an HDR-deprived pathway (text + SDR only), maximizing their KL divergence to enforce HDR grounding and applying dual-entropy regularization to prevent reward hacking. Group-wise rewards include MOS/attribute accuracy, reasoning quality, and self-rewarding. Right: a LoRA-tuned LLM decodes the HDR-aware reasoning; visual inputs originate from both a standard encoder and our HDR-aware adapter.}
\label{fig:architecture}
\end{figure*}


\section{Preliminaries}
\label{sec:prelim}

\label{sec:prelim:grpo}
Large-scale multimodal reinforcement learning requires stable optimization without the high variance of critic-based methods such as PPO~\citep{schulman2017proximal}. Group Relative Policy Optimization (GRPO)~\citep{grpo} achieves this by normalizing rewards within a sampled response group, eliminating the need for a learned value network while preserving sample efficiency. GRPO~\citep{grpo} has become a key component of modern LLM and MLLM post-training pipelines~\citep{yu2025dapo, chu2025sft}, particularly when direct reward modeling is infeasible.

\noindent
\textbf{Formulation.}  
Given a multimodal dataset $\mathcal{D}=\{(q,I,a)\}$ with input prompt $q$, multimodal input $I$, and target answer $a$, we sample $K$ candidate completions $\{o_i\}_{i=1}^{K}$ from the previous policy $\pi_{\theta_{\text{old}}}$. Each completion receives a scalar reward $R_i$, and its normalized group-relative advantage is computed as:

\begin{equation}
\begin{aligned}
\hat{A}_i &= 
\frac{R_i - \mu_R}{\sigma_R + \epsilon}, \quad \mu_R = \frac{1}{K}\sum_{j=1}^{K} R_j, \\ 
\quad
\sigma_R &= \sqrt{\frac{1}{K}\sum_{j=1}^{K}(R_j - \mu_R)^2 }.
\end{aligned}
\label{eq:grpoadv}
\end{equation}

The clipped surrogate objective becomes:
\begin{equation}
\begin{aligned}
\mathcal{J}_{\mathrm{GRPO}}(\theta)
&= \mathbb{E}_{(q,I)\sim\mathcal{D},\,o_i\sim\pi_{\theta_{\mathrm{old}}}}
\frac{1}{K}\sum_{i=1}^{K}\frac{1}{|o_i|}\sum_{t} 
\Big[
\min\!\big(\rho_{i,t}\hat A_i,\,
\mathrm{clip}(\rho_{i,t},1-\epsilon,1+\epsilon)\hat A_i\big) 
-\beta\,D_{\mathrm{KL}}(\pi_\theta\|\pi_{\mathrm{ref}}) \Big].
\end{aligned}
\label{eq:grpo_obj}
\end{equation}

where $\rho_{i,t}=\pi_\theta(o_{i,t}|q,I,o_{i,<t})/\pi_{\theta_{\mathrm{old}}}(o_{i,t}|q,I,o_{i,<t})$ denotes the token-level importance ratio, and $\pi_{\mathrm{ref}}$ is a frozen reference policy that anchors stability.


\noindent
\textbf{Limitations.}  
Despite its stability, vanilla GRPO~\citep{grpo} lacks explicit mechanisms to ensure that the learned policy grounds its behavior in perceptual cues from the input modality~\citep{, papo}. In perception-heavy tasks such as HDR-UGC VQA, this leads to modality neglect~\citep{modalityneglect, papo}, where the policy achieves high textual coherence yet ignores HDR visual information. It also treats all output tokens equally, disregarding token-level uncertainty and reasoning structure issues critical in multimodal reasoning tasks. These limitations motivate our proposed \textcolor{hdrblue}{HDR-Aware Policy Optimization (HAPO)} (Sec.~\ref{sec:hapo}), which extends GRPO~\citep{grpo} with HDR–SDR contrastive grounding, dual-entropy regularization, and entropy-weighted advantage shaping.

\section{Method: \textcolor{hdrblue}{HDR-Q}}
\label{sec:method}

We introduce \textbf{\textcolor{hdrblue}{HDR-Q}}, a multimodal large language model (MLLM) designed for perceptual quality assessment of HDR user-generated videos. The framework couples an HDR-aware vision encoder with a reinforcement learning (RL) objective, HDR-Aware Policy Optimization (HAPO), that explicitly enforces HDR grounding, stabilizes learning against reward hacking, and improves reasoning fidelity. As illustrated in Fig.~\ref{fig:architecture}, HDR-Q integrates both perceptual and reasoning pathways: (i) the HDR-aware encoder yields HDR-sensitive embeddings that capture luminance extremes and color-volume fidelity, while (ii) HAPO fine-tunes the policy to rely on these cues through contrastive, entropy-regularized RL.


\noindent
\subsection{HDR-Aware Vision Encoder}
\label{sec:adapter}
Let $v=\{x_t\}_{t=1}^{T}$ denote a 10-bit HDR video in PQ (BT.2020). We preserve the HDR signal at full precision avoiding tone compression to retain near-black structure, highlight dynamics, and wide-gamut color relationships. For contrastive supervision, an SDR counterpart $v^{SDR}=\{TM(x_t)\}_{t=1}^{T}$ is obtained via a deterministic tone-mapping operator $TM(\cdot)$ (PQ$\!\to\!\gamma$ mapping, quantization, and BT.709 contraction). 

\begin{wrapfigure}{r}{0.5\linewidth}
\vspace{-5pt}
\centering
\includegraphics[width=\linewidth]{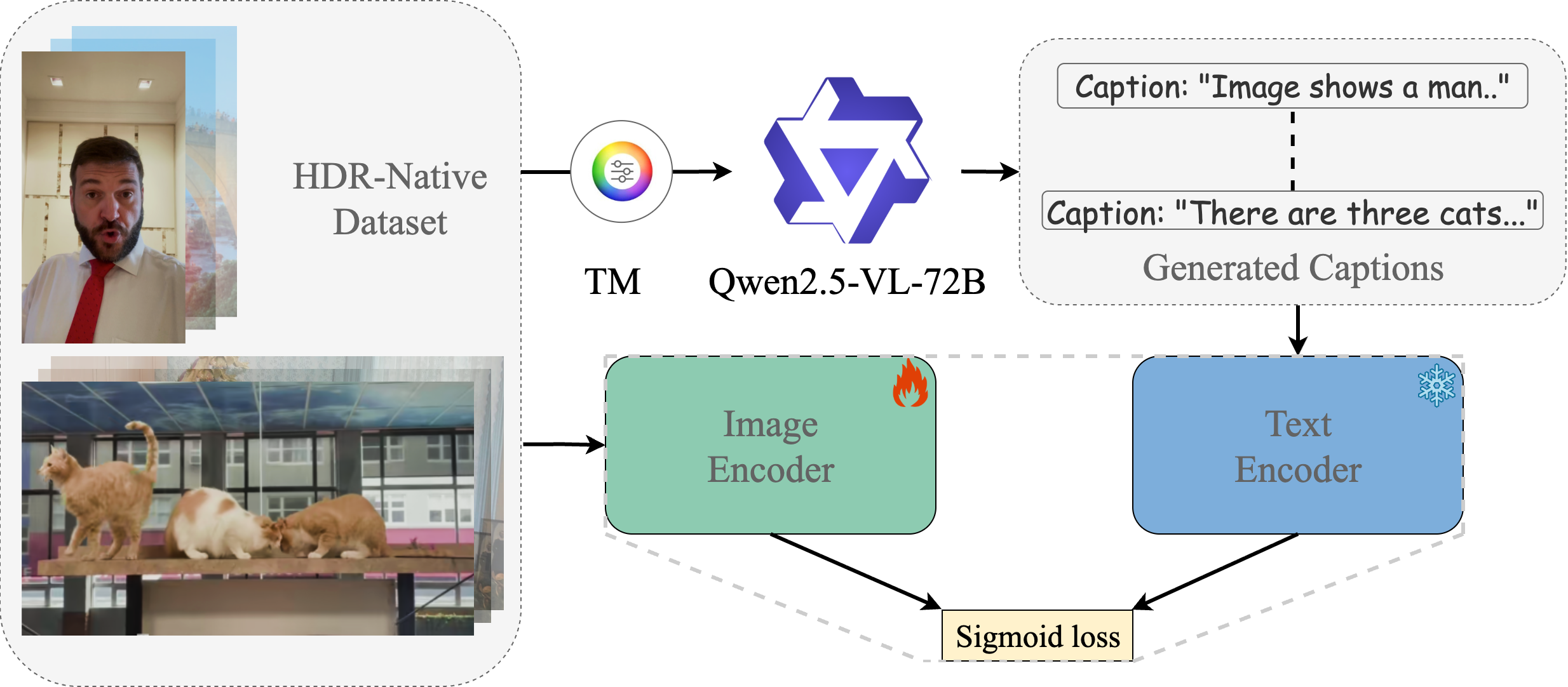}
\caption{\textbf{HDR-aware vision encoder finetuning.}
We adapt SigLIP-2~\citep{siglip2} using HDR–SDR frame–caption pairs with captions generated by Qwen2.5-VL-72B, promoting perceptually aligned HDR embeddings.}
\label{fig:siglip-train}
\vspace{-5pt}
\end{wrapfigure}

We adapt a pretrained SigLIP-2 encoder~\citep{siglip2} $\mathcal{E}_{\psi}$ on HDR frame–caption pairs $(x_t,c_t)$, where captions are generated by Qwen2.5-VL-72B~\citep{qwen}. The goal is to yield embeddings that remain semantically aligned yet intrinsically sensitive to HDR variations. 

\noindent
\paragraph{Dual-Domain Supervision.}
A key challenge is that generic captions $c_t$ are equally valid for $x_t$ and $x^{SDR}_t$, potentially causing collapse where HDR and SDR embeddings overlap. To avoid this, we introduce dual-domain supervision. For each HDR frame $x_t$, we generate $x^{SDR}_t$ and enforce contrastive separation: the HDR embedding must remain closer to its caption than the SDR embedding:

\begin{equation}
\begin{aligned}
\mathcal{L}_{\text{contrast}}
&= \max\!\Big(
0,\;
\delta - D\!\big(\mathcal{E}_{\psi}(x_t),\,\mathcal{E}_{\psi}(c_t)\big)
+\, D\!\big(\mathcal{E}_{\psi}(x^{SDR}_t),\,\mathcal{E}_{\psi}(c_t)\big)
\Big).
\end{aligned}
\end{equation}

where $D(\cdot,\cdot)$ denotes cosine distance and $\delta$ is a margin.  
The full encoder loss combines alignment and HDR discrimination:

\begin{equation}
\mathcal{L}_{\mathrm{enc}} = \mathcal{L}_{\mathrm{Sigmoid}}(x_t,c_t) 
+ \lambda_{\mathrm{ctr}}\mathcal{L}_{\mathrm{contrast}}
\end{equation}

ensuring that the learned embeddings remain semantically faithful while being perceptually attuned to HDR contrast and luminance cues (see Fig.~\ref{fig:siglip-train}).


\subsection{HDR-Aware Policy Optimization (HAPO)}
\label{sec:hapo}

While GRPO~\citep{grpo} stabilizes multimodal RL, it offers no guarantee that the policy exploits visual cues rather than textual priors. HAPO extends GRPO~\citep{grpo} with three HDR-specific components that explicitly enforce modality grounding (See Appendix \textcolor{red}{E}):
\noindent
\paragraph{(i) HDR–SDR Contrastive KL.}
To prevent modality neglect~\citep{modalityneglect}, we contrast rollouts with and without HDR tokens:
\begin{equation}
\small
\mathcal{K}_{\text{HDR}}(\theta) =
D_{\text{KL}}\!\big(\pi_\theta^{HDR}\,\|\,\pi_\theta^{SDR}\big)
\label{eq:klhdr}
\end{equation}
where $\pi_\theta^{HDR}$ and $\pi_\theta^{SDR}$ are policies with and without HDR input. Maximizing $\mathcal{K}_{\mathrm{HDR}}$ ensures that removing HDR tokens significantly perturbs the decoding distribution, thereby incentivizing the model to exploit HDR-specific information rather than collapsing into SDR-only reasoning. 

\noindent
\paragraph{(ii) Dual-Entropy Regularization.}
A well-known pitfall in contrastive KL maximization is entropy inflation, the policy can trivially satisfy the objective by producing overly uncertain outputs~\citep{rafailov2023direct, zeng2024uncertainty}. To prevent this, we introduce policy entropy regularization on both HDR and SDR pathways:

\begin{equation}
\begin{aligned}
\mathcal{H}_{\text{dual}}(\theta)
&= \mathbb{E}_{o\sim\pi_{\theta_{old}}}
\frac{1}{K}\!\sum_{i,t} \Big[
\eta_1\,\mathcal{H}\!\left(\pi_\theta^{HDR}(o_{i,t})\right) 
+\, \eta_2\,\mathcal{H}\!\left(\pi_\theta^{SDR}(o_{i,t})\right)
\Big].
\end{aligned}
\end{equation}

where $\mathcal{H}$ denotes token-level entropy, i.e. $\mathcal{H}(\pi_{\theta})=\log \pi_{\theta}$, and $\eta_1$ and $\eta_2$ are hyperparameters. This prevents collapse while preserving sharp, HDR-grounded distributions.

\noindent
\paragraph{(iii) High-Entropy Weighting (HEW).}
GRPO assigns the same normalized advantage $\hat A_i$ to all tokens of a completion $o_i$, regardless of their informativeness. However, recent work~\citep{cui2505entropy} demonstrates that reinforcement learning benefits from focusing policy gradients of tokens promoting exploration, and thus improving reasoning, while tokens following fixed reasoning path provide little signal. In HDR-UGC VQA, high-entropy tokens typically occur when the model must identify or calibrate HDR-specific distortions (e.g., banding in gradients, highlight clipping, near-black crushing). By amplifying the learning signal at these tokens, HEW directs policy optimization toward the most informative reasoning steps, yielding stronger HDR grounding and more precise MOS predictions. We then rescale the group-normalized advantage $\hat A_i$ into a token-specific advantage:
\begin{equation}
\begin{aligned}
w_{i,t}
&= \mathrm{clip}\!\Bigg(
1+\lambda_{\mathrm{HEW}}
\frac{H_{i,t}}{\tfrac{1}{|o_i|}\sum_{t'=1}^{|o_i|} H_{i,t'}},
\, w_{\min},\, w_{\max}
\Bigg),
\quad
\tilde{A}_{i,t} &= w_{i,t}\cdot \hat A_i .
\end{aligned}
\label{eq:hew}
\end{equation}
where $H_{i,t}$ is per-token entropy.

\noindent
\paragraph{Full HAPO Objective.}
Combining these terms yields:
\begin{equation}
\begin{aligned}
\mathcal{J}_{\text{HAPO}}(\theta)
= \mathbb{E}_{o\sim\pi_{\theta_{\text{old}}}}\Bigg[
\frac{1}{K}\sum_{i,t}
\min\!\Big(
\rho_{i,t}\tilde{A}_{i,t},\,
\mathrm{clip}(\rho_{i,t},1-\epsilon,1+\epsilon)\tilde{A}_{i,t}
\Big)
\Bigg] \\
\quad
-\beta\,D_{\text{KL}}\!\left(\pi_\theta^{HDR}\,\|\,\pi_{\text{ref}}\right)
+\gamma\,\mathcal{K}_{\text{HDR}}(\theta)
-\mathcal{H}_{\text{dual}}(\theta).
\end{aligned}
\end{equation}

This enforces HDR-aware reasoning while maintaining stable optimization.

\noindent
\paragraph{Mutual Information Perspective.}
Our HDR–SDR contrastive KL can be interpreted as enforcing an information-theoretic dependency between HDR inputs and model outputs. Let $v$ denote the HDR video, $v^{SDR}$ its SDR tone-mapped counterpart, and $o$ the output sequence. By applying variational mutual information bounds~\citep{ishmael2018mine, ma2023mutual}, we obtain $\mathbb{E}_{v,v^{SDR}}\!\left[\mathcal{K}_{\mathrm{HDR}}(\theta)\right]$ as:
\begin{equation}
\begin{aligned}
\mathbb{E}_{v,v^{SDR}}\,
\mathbb{E}_{o\sim \pi_\theta(\cdot\,|\,v,v^{SDR})}
\Bigg[
\log
\frac{
\pi_\theta(o\,|\,v,v^{SDR})
}{
\pi_\theta(o\,|\,v^{SDR})
}
\Bigg]
\;\;\ge\;
I_\theta\!\left(o;\,v,v^{SDR}\,\middle|\,v^{SDR}\right)
-\kappa_\theta .
\end{aligned}
\label{eq:info_bound}
\end{equation}

where $I_\theta(o;v,v^{SDR}|v^{SDR})$ is the conditional mutual information under $\pi_\theta$, and $\kappa_\theta$ captures mismatch due to conditioning on $v^{SDR}$. This result shows that maximizing~\eqref{eq:klhdr} provably increases HDR informativeness, ensuring the policy relies on HDR-specific cues rather than collapsing to SDR-only reasoning.

\subsection{Rewards and Training Pipeline}
\label{sec:rewards_training}
HAPO jointly optimizes three reward signals: format ($R_{\text{fmt}}$), regression accuracy ($R_{\text{sc}}$)~\citep{qinsight,depictqa,visualquality}, and self-consistency ($R_{\text{self}}$)~\citep{selfreward} combined as
\begin{equation}
\mathcal{R}_i=w_{\text{fmt}}R_{\text{fmt}}+
w_{\text{sc}}R_{\text{sc}}+
w_{\text{self}}R_{\text{self}}.
\end{equation}
A Gaussian-weighted score reward stabilizes fine-grained MOS prediction, while the self-reward consolidates within-group consensus.

\noindent
\paragraph{Two-Stage RL Training.}
Our training follows a two-stage RL-based paradigm~\citep{sft-rl, qoq-med}, both optimized with the same objective but serving distinct purposes:
\begin{itemize}
\item \textbf{Stage 1 (Modality Alignment):} aligns HDR tokens and projection layers via short HAPO runs.
\item \textbf{Stage 2 (Full-RFT):} applies complete HAPO optimization on the HDR-UGC corpus, balancing distortion diversity and reasoning quality.
\end{itemize}
Overall, \textbf{\textcolor{hdrblue}{HDR-Q}} unifies perceptual sensitivity and reasoning stability, the HDR encoder injects physical luminance awareness, contrastive KL enforces grounding, entropy regularization curbs uncertainty, and HEW refines token-level learning yielding accurate, interpretable HDR-aware quality judgments.


\definecolor{hdrblue}{RGB}{46,105,173}
\definecolor{hdrgold}{RGB}{214,170,0}
\definecolor{hdrgray}{RGB}{245,247,250}
\definecolor{hdrred}{RGB}{190,50,50}
\definecolor{hdrgreen}{RGB}{30,145,90}

\begin{table}[htbp]
\centering
\small
\caption{Performance on \textbf{Beyond8Bits}. Best results in \textcolor{hdrblue}{\textbf{blue bold}}, second best are \underline{underlined}}.
\label{tab:results-1}
\renewcommand{\arraystretch}{0.8}
\resizebox{0.95\linewidth}{!}{%
\begin{tabular}{lcccc}
\toprule
\textbf{Model} & SRCC(\(\uparrow\)) & PLCC(\(\uparrow\)) & RMSE(\(\downarrow\)) & KRCC(\(\uparrow\)) \\
\midrule
\rowcolor{hdrgold!10}
\multicolumn{5}{c}{\textbf{DL models}} \\
BRISQUE~\citep{BRISQUE}       & 0.4096 & 0.4689 & 11.7019 & 0.2797 \\
CONTRIQUE~\citep{CONTRIQUE}     & 0.6245 & 0.6054 & 15.0224 & 0.4464 \\
RE-IQA~\citep{REIQA}        & 0.5698 & 0.5441 & 17.9049 & 0.4038 \\
VBLIINDS~\citep{V-BLINDS}      & 0.4440 & 0.4397 & 11.7234 & 0.3044 \\
CONVIQT~\citep{conviqt}       & 0.7987 & 0.8099 &  8.4807 & 0.6095 \\
FastVQA~\citep{FASTVQA}       & 0.4909 & 0.4193 & 26.1325 & 0.3398 \\
FasterVQA~\citep{FASTERVQA}     & 0.4808 & 0.3224 & 29.6357 & 0.3367 \\
DOVER~\citep{Dover}         & 0.5094 & 0.5037 & 16.7176 & 0.3548 \\
COVER~\citep{cover}         & 0.6645 & 0.6645 & 16.8597 & 0.4870 \\
HDRMAX~\citep{LIVE-HDR}        & 0.6054 & 0.6070 & 10.1400 & 0.4277 \\
HDRChipQA~\citep{HDR-ChipQA}     & 0.7180 & 0.7290 &  8.2987 & 0.5282 \\
HIDROVQA~\citep{HIDRO-VQA}      & 0.8508 & 0.8784 &  \underline{6.0875} & 0.6694 \\
\midrule
\rowcolor{hdrgold!10}
\multicolumn{5}{c}{\textbf{MLLM base model}} \\
Qwen2.5-VL(7B)~\citep{qwen}       & 0.3089 & 0.3228 & 27.8899 & 0.2432 \\
GLM-4.1V-Thinking(9B)~\citep{glm}      & 0.2641 & 0.3944 & 23.8883 & 0.2924 \\
Ovis2.5(9B)~\citep{ovis}       & 0.3423 & 0.3860 & 26.7570 & 0.2823 \\
OmniLong-Qwen2.5-VL(7B)~\citep{OmniLong}     & 0.3472 & 0.3595 & 25.7616 & 0.2677 \\
\midrule
\rowcolor{hdrgold!10}
\multicolumn{5}{c}{\textbf{MLLM VQA model}} \\
Q-Align~\citep{qalign}       & 0.4615 & 0.3673 & 20.3411 & 0.3257 \\
Q-Insight~\citep{qinsight}     & 0.5170 & 0.5621 & 20.7832 & 0.4138 \\
Q-Instruct~\citep{qinsight}    & 0.5035 & 0.4712 & 19.6567 & 0.3496 \\
DeQA~\citep{deqa}          & 0.5064 & 0.4642 & 19.5772 & 0.3586 \\
Visual-Quality-Q1~\citep{visualquality} & 0.3909 & 0.3617 &  23.6462 & 0.2809 \\
\midrule
HDR-Q (SDR) & \underline{0.8914} & \underline{0.8895} & 7.4240 & \underline{0.7052} \\
\rowcolor{hdrblue!20}
\textbf{\textcolor{hdrblue}{HDR-Q (full)}} & \textcolor{hdrblue}{\textbf{0.9206}} & \textcolor{hdrblue}{\textbf{0.9118}} & \textcolor{hdrblue}{\textbf{5.1594}} & \textcolor{hdrblue}{\textbf{0.7218}} \\
\bottomrule
\end{tabular}}
\end{table}



\section{Experiments}
\label{sec:experiments}

\subsection{Experimental Setup}
\label{sec:exp_setup}

\vspace{-1.5em}
\noindent
\paragraph{Datasets.}
We evaluate \textcolor{hdrblue}{\textbf{HDR-Q}} on the curated \textbf{Beyond8Bits} benchmark and test generalization on two public HDR-VQA datasets: LIVE-HDR~\citep{LIVE-HDR} and SFV+HDR~\citep{sfv+hdr}. \textbf{Beyond8Bits} is split by source identity into 70\%/20\%/10\% train/val/test to avoid overlap.

\vspace{-1.5em}
\noindent
\paragraph{Metrics.}
Following VQA convention~\citep{conviqt,KVQ,chug,HIDRO-VQA}, we report Spearman’s Rank (SRCC), Pearson’s Linear (PLCC), and Kendall’s Rank (KRCC) correlations (\(\uparrow\) higher is better), and RMSE (\(\downarrow\) lower is better) against MOS.

\vspace{-1.5em}
\noindent
\paragraph{Baselines.}
We compare four category of methods.  
(i) NR-VQA: BRISQUE~\citep{BRISQUE}, VBLIINDS~\citep{V-BLINDS}, FastVQA~\citep{pamifastvqa}, FasterVQA~\citep{FASTERVQA}, DOVER~\citep{Dover}, CONVIQT~\citep{conviqt}, COVER~\citep{cover};  
(ii) HDR-VQA: HDRMAX~\citep{hdrmax-1}, HDR-ChipQA~\citep{HDR-ChipQA}, HIDRO-VQA~\citep{HIDRO-VQA};  
(iii) MLLM/VLM-VQA: Q-Align~\citep{qalign}, Q-Instruct~\citep{qinstruct}, Q-Insight~\citep{qinsight}, DeQA~\citep{deqa}, Visual-Quality-R1~\citep{visualquality};  
(iv) Base MLLMs: Qwen2.5-VL~\citep{qwen}, GLM-4.1V-Thinking~\citep{glm}, Ovis2.5~\citep{ovis}, OmniLong-Qwen2.5-VL~\citep{OmniLong}.  
Where applicable, methods are re-trained on \textbf{Beyond8Bits} using authors’ protocols; others are evaluated in their released form.

\begin{wrapfigure}{r}{0.55\linewidth}
\centering
\includegraphics[width=\linewidth]{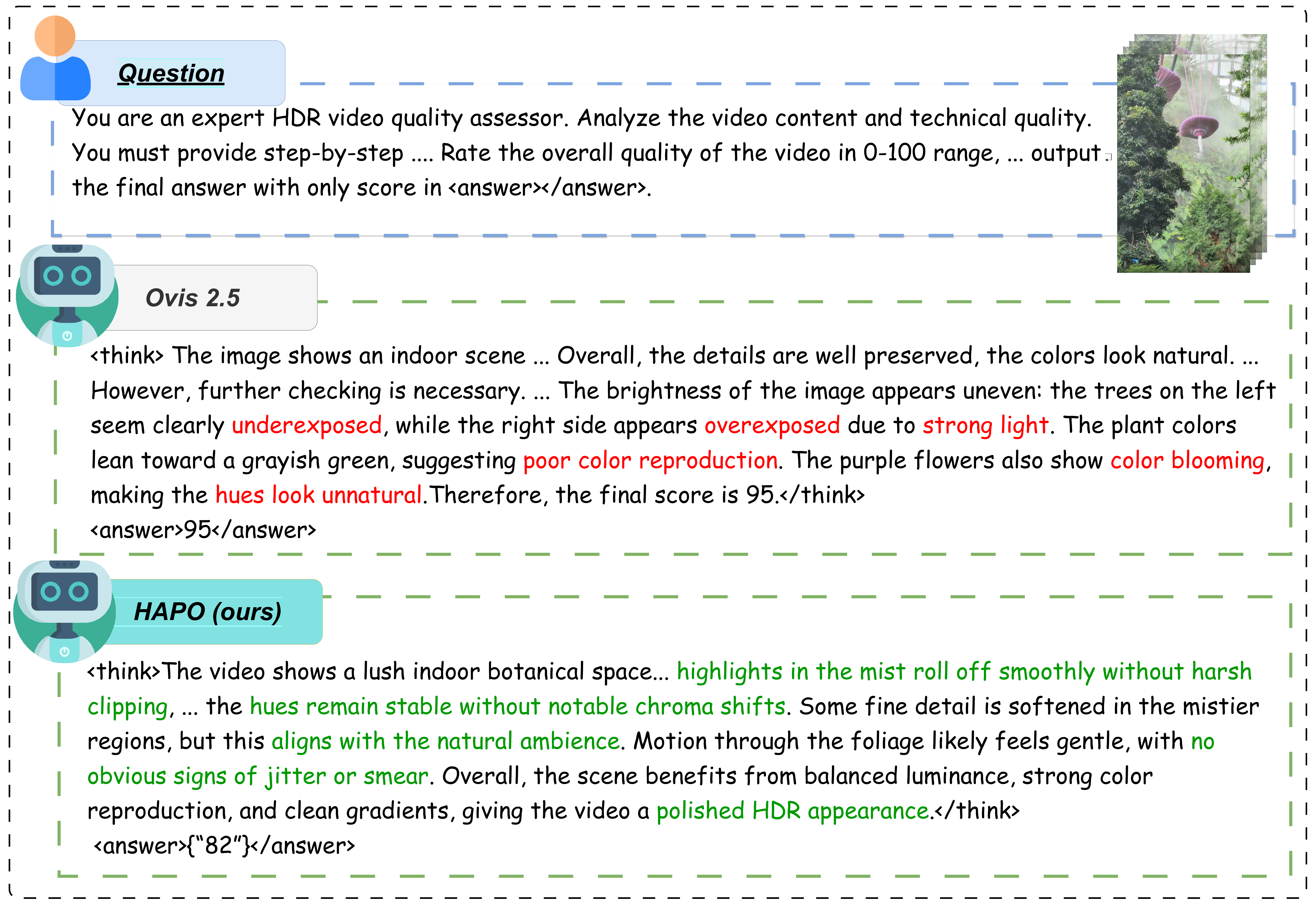}
\caption{Given the same HDR video, OVIS 2.5 produces multiple incorrect judgments. In contrast, our HAPO-enhanced \textcolor{hdrblue}{\textbf{HDR-Q}} provides HDR-grounded reasoning. (Best viewed zoomed in)}
\label{fig:example-reasone}
\vspace{-8pt}
\end{wrapfigure}

\vspace{-1.5em}
\noindent
\paragraph{Implementation details.}
\textcolor{hdrblue}{\textbf{HDR-Q}} is built on Ovis2.5~\citep{ovis} with rank-4 LoRA adapters~\citep{lora}. Frames are ingested at native 10-bit PQ (no linear downscaling). Each clip is uniformly sampled into $T=8$ frames; visual tokens from $\mathcal{E}_{\psi}(x_t)$ and SDR tokens from $\mathcal{E}_{\psi}(x^{\mathrm{SDR}}_t)$ feed the language decoder via learned projections. In HAPO, group size $K=8$; clip range $\epsilon=0.1$ (clip-higher); reference KL weight $\beta=0.02$; HDR–SDR contrastive KL weight $\gamma=0.5$; policy entropy $\eta_1,\eta_2=0.01,0.05$; HEW modulation $\lambda_{\mathrm{HEW}}=0.3$ with $w_{\min}=0.5,w_{\max}=2.0$. In gaussian score reward $R_{\mathrm{sc}}$ we use $\sigma=3$ and $\alpha=1$; weights $(w_{\mathrm{fmt}},w_{\mathrm{sc}},w_{\mathrm{self}})$ tuned on validation. 
We use AdamW~\citep{adamw}, lr $1\!\times\!10^{-5}$, batch size of $4$. We use four NVIDIA H200 GPUs for training.

\begin{table*}[htbp]
\centering
\small
\caption{Cross-dataset performance Comparison on LIVE-HDR~\citep{LIVE-HDR} and SFV+HDR~\citep{sfv+hdr} Datasets.} 
\label{tab:sota-other}
\renewcommand{\arraystretch}{0.8}
\resizebox{\linewidth}{!}{%
\begin{tabular}{l c c c c || c c c c}
\toprule
\textbf{Model}
  & \multicolumn{4}{c||}{\textbf{LIVE-HDR}} 
  & \multicolumn{4}{c}{\textbf{SFV+HDR}} \\
\cmidrule(lr){2-9}
  & SROCC(\(\uparrow\)) & PLCC(\(\uparrow\)) & RMSE(\(\downarrow\)) & KRCC(\(\uparrow\)) 
  & SROCC(\(\uparrow\)) & PLCC(\(\uparrow\)) & RMSE(\(\downarrow\)) & KRCC(\(\uparrow\)) \\
\midrule
\rowcolor{hdrgold!10}
\multicolumn{9}{c}{\textbf{DL models}} \\
BRISQUE~\citep{BRISQUE}     & 0.7251 & 0.7139 & 12.6404 & 0.3424 & 0.4664 & 0.4186 & 0.3811 & 0.3165 \\
CONTRIQUE~\citep{CONTRIQUE} & 0.8170 & 0.7875 & 11.2514 & 0.5876 & 0.5901 & 0.5959 & 0.3368 & 0.4204 \\
RE-IQA~\citep{REIQA}         & 0.7196 & 0.6883 & 15.1653 & 0.5197 & 0.5822 & 0.5998 & 0.3072 & 0.4145 \\
VBLIINDS~\citep{V-BLINDS}   & 0.7483 & 0.7193 & 12.7794 & 0.2541 & 0.3335 & 0.2713 & 0.3988 & 0.2300 \\
CONVIQT~\citep{conviqt}     & 0.7922 & 0.8001 & 11.9681 & 0.6041 & 0.5736 & 0.6017 & 0.3412 & 0.4170 \\
FastVQA~\citep{FASTVQA}     & 0.5182 & 0.5727 & 18.8379 & 0.3822 & 0.7130 & 0.7295 & 0.7467 & 0.5193 \\
FasterVQA~\citep{pamifastvqa}& 0.3385 & 0.4114 & 22.1425 & 0.2282 & 0.6948 & 0.6889 & 0.3081 & 0.5089 \\
DOVER~\citep{Dover}         & 0.6303 & 0.6832 & 17.0005 & 0.4692 & 0.6001 & 0.6154 & 0.5750 & 0.4270 \\
COVER~\citep{cover}         & 0.5022 & 0.5013 & 21.3297 & 0.3731 & 0.6613 & 0.7048 & 0.6831 & 0.4705 \\
HDRMAX~\citep{LIVE-HDR}     & 0.6308 & 0.5088 & 15.4146 & 0.4509 & 0.5371 & 0.5463 & 0.3495 & 0.3821 \\
HDRChipQA~\citep{HDR-ChipQA}& 0.8250 & 0.8344 &  9.8038 & 0.4501 & 0.6296 & 0.6508 & 0.3271 & 0.4440 \\
HIDROVQA~\citep{HIDRO-VQA}  & \underline{0.8793} & \underline{0.8678} &  \underline{8.8743} & \underline{0.6919} & \underline{0.7003} & \underline{0.7320} & \underline{0.2735} & \underline{0.5156} \\
\midrule
\rowcolor{hdrgold!10}
\multicolumn{9}{c}{\textbf{MLLM base model}} \\
Qwen2.5-VL(7B)~\citep{qwen}         & 0.3099 & 0.3630 & 30.2082 & 0.2411 & 0.2925 & 0.2696 & 0.7480 & 0.2270 \\
GLM-4.1V-Thinking(9B)~\citep{glm}         & 0.4513 & 0.5517 & 26.3800 & 0.3464 & 0.5971 & 0.6066 & 0.4591 & 0.4484 \\
Ovis2.5(9B)~\citep{ovis}         & 0.2948 & 0.3124 & 29.7789 & 0.2154 & 0.5909 & 0.5317 & 0.7016 & 0.4528 \\
OmniLong-Qwen2.5-VL(7B)~\citep{OmniLong}       & 0.2403 & 0.2223 & 29.7394 & 0.1853 & 0.2363 & 0.2212 & 0.7403 & 0.1823 \\
\midrule
\rowcolor{hdrgold!10}
\multicolumn{9}{c}{\textbf{MLLM VQA model}} \\
Q-Align~\citep{qalign}        & 0.3346 & 0.3604 & 19.8287 & 0.2313 & 0.6968 & 0.6709 & 0.5097 & 0.4991 \\
Q-Insight~\citep{qinsight}       & 0.3675 & 0.3825 & 25.0578 & 0.2820 & 0.6266 & 0.4685 & 0.6636 & 0.4747 \\
Q-Instruct~\citep{qinstruct}      & 0.4083 & 0.4340 & 23.1015 & 0.2839 & 0.5830 & 0.5501 & 1.0250 & 0.3975 \\
DeQA~\citep{deqa}            & 0.3321 & 0.3809 & 19.3193 & 0.2298 & 0.6850 & 0.6721 & 0.4452 & 0.4845 \\
Visual-Quality-Q1~\citep{visualquality} & 0.4824 & 0.5394 & 20.8971 & 0.3564 & 0.5955 & 0.5577 & 0.5878 & 0.4416 \\
\midrule
HDR-Q (SDR) & 0.8542 & 0.8445 & 12.4121 & 0.6681 & 0.6971 & 0.7019 & 0.3075 & 0.4885 \\
\rowcolor{hdrblue!20}
\textbf{\textcolor{hdrblue}{HDR-Q (full)}} & \textcolor{hdrblue}{\textbf{0.9081}} & \textcolor{hdrblue}{\textbf{0.8978}} & \textcolor{hdrblue}{\textbf{7.6031}} & \textcolor{hdrblue}{\textbf{0.7363}} & \textcolor{hdrblue}{\textbf{0.7251}} & \textcolor{hdrblue}{\textbf{0.7502}} & \textcolor{hdrblue}{\textbf{0.2514}} & \textcolor{hdrblue}{\textbf{0.5261}} \\
\bottomrule
\end{tabular}}
\end{table*}

\begin{table*}[h!]
\centering
\caption{\textbf{Component ablation on \textbf{Beyond8Bits}.} \cmark=enabled, \xmark=disabled. 
CoT len: CoT length and Tok. H: mean token entropy.}
\label{tab:ablate_all}
\renewcommand{\arraystretch}{1}
\resizebox{\textwidth}{!}{
\begin{tabular}{lcccccc|cccc|cc}
\toprule
\textbf{Variant} & \textbf{HDR-Enc.} & \textbf{HAPO} & \textbf{HDR--SDR KL} & \textbf{Dual Ent.} & \textbf{HEW} & \textbf{Self-R.} 
& \textbf{PLCC} & \textbf{SRCC} & \textbf{RMSE} & \textbf{KRCC} & \textbf{CoT len} & \textbf{Tok. $H$} \\
\midrule
\rowcolor{gray!15}
GRPO baseline & \xmark & \xmark & \xmark & \xmark & \xmark & \xmark 
& 0.79 & 0.81 & 10.73 & 0.56 & 168 & 0.20\\

\rowcolor{hdrblue!2}
GRPO + HDR-Enc. & \cmark & \xmark & \xmark & \xmark & \xmark & \xmark 
& 0.81 & 0.83 & 8.96 & 0.61 & 161 & 0.24\\

\rowcolor{hdrblue!4}
HAPO w/o HDR--SDR KL & \cmark & \cmark & \xmark & \cmark & \cmark & \cmark 
& 0.84 & 0.86 & 7.10 & 0.64 & 142 & 0.29\\

\rowcolor{hdrblue!6}
HAPO w/o Dual Ent. & \cmark & \cmark & \cmark & \xmark & \cmark & \cmark 
& 0.89 & 0.91 & 5.82 & 0.71 & 148 & 0.26\\

\rowcolor{hdrblue!8}
HAPO w/o HEW & \cmark & \cmark & \cmark & \cmark & \xmark & \cmark 
& 0.87 & 0.88 & 6.11 & 0.68 & 155 & 0.27\\

\rowcolor{hdrblue!10}
HAPO w/o Self-Reward & \cmark & \cmark & \cmark & \cmark & \cmark & \xmark 
& \underline{0.90} & \underline{0.92} & \underline{5.22} & \underline{0.71} & 140 & 0.31\\

\rowcolor{hdrblue!20}
\textbf{\textcolor{hdrblue}{HDR-Q (Full)}} & \textbf{\cmark} & \textbf{\cmark} & \textbf{\cmark} & \textbf{\cmark} & \textbf{\cmark} & \textbf{\cmark} 
& \textcolor{hdrblue}{\textbf{0.91}} & \textcolor{hdrblue}{\textbf{0.92}} & \textcolor{hdrblue}{\textbf{5.15}} & \textcolor{hdrblue}{\textbf{0.72}} 
& \textcolor{hdrblue}{\textbf{137}} & \textcolor{hdrblue}{\textbf{0.33}}\\
\bottomrule
\end{tabular}}
\end{table*}

\subsection{Main Results}
\label{sec:main_results}
Table~\ref{tab:results-1} reports quantitative results on \textbf{Beyond8Bits}. \textcolor{hdrblue}{\textbf{HDR-Q}} consistently outperforms all SDR, HDR, and MLLM-based baselines across correlation metrics, with substantial gains in RMSE. Against HDR-ChipQA~\citep{HDR-ChipQA} and HIDRO-VQA~\citep{HIDRO-VQA}, \textcolor{hdrblue}{\textbf{HDR-Q}} achieves higher SRCC/PLCC with lower RMSE, indicating the benefits of HDR-aware embeddings plus HAPO grounding. Against FastVQA~\citep{pamifastvqa} and DOVER~\citep{Dover}, \textcolor{hdrblue}{\textbf{HDR-Q}} remains robust despite diverse UGC capture pipelines. Relative to all MLLM and VLM models, \textcolor{hdrblue}{\textbf{HDR-Q}}’s gains stem from HDR-aware encoder finetuning on 10-bit PQ without linear SDR scaling, and HDR–SDR contrastive KL (prevents modality neglect).

\begin{wrapfigure}{r}{0.55\linewidth}
\vspace{-7pt}
\centering
\begin{subfigure}[b]{0.48\linewidth}
    \centering
    \includegraphics[width=\linewidth]{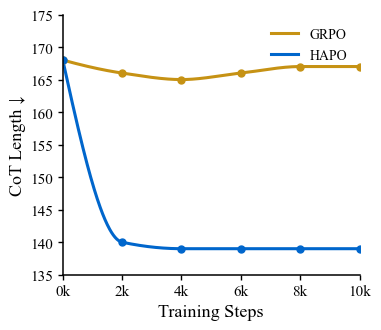}
    \caption{CoT Length}
    \label{fig:cot}
\end{subfigure}
\hfill
\begin{subfigure}[b]{0.48\linewidth}
    \centering
    \includegraphics[width=\linewidth]{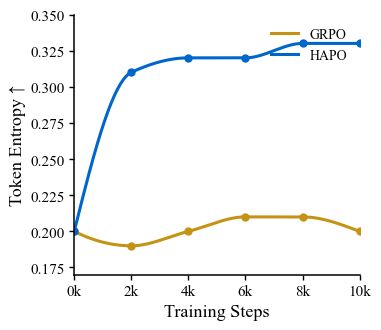}
    \caption{Token Entropy}
    \label{fig:entropy}
\end{subfigure}
\caption{Analysis of Chain-of-Thought (CoT) Length and Token Entropy over training iterations. (a) shows the decrease in CoT length, while (b) shows the corresponding increase in token entropy.}
\label{fig:len_entropy}
\vspace{-10pt}
\end{wrapfigure}

\noindent
To test robustness, we evaluate zero-shot transfer on LIVE-HDR~\citep{LIVE-HDR} and SFV+HDR~\citep{sfv+hdr} (Table~\ref{tab:sota-other}). \textcolor{hdrblue}{\textbf{HDR-Q}} retains high correlation and low error without retraining evidence that its HDR-aware encoder and HAPO grounding produce representations that generalize across UGC and PGC HDR domains.

\noindent
\textcolor{hdrblue}{\textbf{HDR-Q}} generates concise, HDR-aware reasoning (Fig.~\ref{fig:example-reasone}), detecting “natural indoor scene,” “possible hues from chroma shifts,” or “jitter” and linking them to perceptual judgments. Fig.~\ref{fig:len_entropy} shows that HAPO stabilizes CoT length while HEW concentrates gradients on informative tokens yielding efficient and interpretable reasoning.





\subsection{Ablation Studies}
\label{sec:ablation}
\noindent
Table~\ref{tab:ablate_all} quantifies the contribution of each component. Removing HDR finetuning drops SRCC markedly, confirming that 10-bit cues are essential. Omitting HDR–SDR KL causes modality neglect, while disabling entropy regularization yields unstable, verbose reasoning. HEW improves token-level credit assignment, and self-rewarding enhances stability on noisy samples.
\noindent
Fig.~\ref{fig:len_entropy} shows that HAPO reduces unnecessary CoT length over time while maintaining or improving accuracy, suggesting better use of visual evidence rather than increase in boilerplate rationales.

\subsection{Complexity and Throughput}
\label{sec:complexity}
HAPO only adds an additional SDR-path forward pass only during training. Inference cost equals a single HDR path decode, maintaining competitive throughput on NVIDIA H200 GPUs.


\section{Conclusion}
\label{sec:conclusion}
We tackled the critical challenge of perceptual quality assessment for the fast-growing domain of HDR user-generated videos. We introduced \textbf{Beyond8Bits}, the largest crowdsourced subjective dataset for real-world HDR content, spanning diverse scenes, devices, and compression settings. We further proposed \textbf{\textcolor{hdrblue}{HDR-Q}}, the first multimodal large language model for HDR quality assessment, combining our novel HDR-aware vision encoder with HDR-Aware Policy Optimization (HAPO), a reinforcement learning framework that enforces HDR–SDR perceptual grounding and stabilizes reasoning via dual-entropy regularization and entropy-weighted credit assignment. HAPO enables accurate, interpretable, and HDR-sensitive quality reasoning. \textcolor{hdrblue}{\textbf{HDR-Q}} achieves state-of-the-art alignment with human opinion scores across \textbf{Beyond8Bits}, LIVE-HDR, and SFV+HDR. By releasing the dataset, we hope to catalyze future research in HDR-aware perception, evaluation, and generative model alignment.

\section{Acknowledgment}
This work was supported by the National Science Foundation AI Institute for Foundations of Machine Learning (IFML) under Grant 2019844. The authors thank the Texas Advanced Computing Center (TACC) at The University of Texas at Austin for providing VISTA compute infrastructure that contributed to the part of research outcomes in this paper.


\bibliographystyle{assets/plainnat}
\bibliography{main}

\clearpage
\newpage


\end{document}